# Closed-loop Feedback Registration for Consecutive Images of Moving Flexible Targets

Rui Ma[1,3], Xian Du[1,2]
[1]Institute for Applied Life Sciences, University of Massachusetts, Amherst, MA 01003, USA
[2]Mechanical & Industrial Engineering Department, University of Massachusetts, Amherst, MA 01003, USA
[3]Electrical & Computer Engineering Department, University of Massachusetts, Amherst, MA 01003, USA

Corresponding author: Xian Du (e-mail: xiandu@umass.edu).

This work is supported in part by the National Science Foundation (CMMI #1916866, CMMI #1942185, and CMMI #1907250). Any opinions, findings, and conclusions or recommendations expressed in this material are those of the authors and do not necessarily reflect the views of the National Science Foundation.

**ABSTRACT** Advancement of imaging techniques enables consecutive image sequences to be acquired for quality monitoring of manufacturing production lines. Registration for these image sequences is essential for in-line pattern inspection and metrology, e.g., in the printing process of flexible electronics. However, conventional image registration algorithms cannot produce accurate results when the images contain many similar and deformable patterns in the manufacturing process. Such a failure originates from a fact that the conventional algorithms only use the spatial and pixel intensity information for registration. Considering the nature of temporal continuity and consecution of the product images, in this paper, we propose a closed-loop feedback registration algorithm for matching and stitching the deformable printed patterns on a moving flexible substrate. The algorithm leverages the temporal and spatial relationships of the consecutive images and the continuity of the image sequence for fast, accurate, and robust point matching. Our experimental results show that our algorithm can find more matching point pairs with a lower root mean squared error (RMSE) compared to other state-of-the-art algorithms while offering significant improvements to running time.

## I. INTRODUCTION

Image registration is a process of transforming two or more images into a single coordinate system. It is a fundamental step in many image processing and computer vision tasks, such as biomedical image diagnosis [1], super-resolution [2], 3-D reconstruction [3], object recognition [4], motion tracking [5], image retrieval [6], and action recognition [7]. Designing and developing an accurate, effective, and robust image registration algorithm is critical to the success of these tasks because the registration results have a significant impact on the quality of the follow-up data processing steps. Inaccurate registration results could detriment all the follow-up steps.

Over the past decades, many image registration algorithms were proposed. These algorithms can be categorized into area-based methods [8]–[10] and feature-based methods [11]–[15].

Area-based methods, sometimes called correlation-like methods, usually take the pre-defined fixed size image patch or even the entire image as the reference and compare the reference to the target image with a specific similarity measurement. The similarity measurements usually contain the summation of absolute values of differences, cross-correlation, mutual information and phase correlation [16]. After locating the most similar position on the target image, the corresponding position pair between the reference and the target is created. However, there are several limitations on the area-based registration methods. First, the image patch, commonly defined as the rectangular window, suits the registration of images which are locally deformed by a translation transformation. If the images are deformed by more complex transformations, the image patch cannot cover the same parts of the scene between the reference and the target images. Second, if the target images contain too many similar and deformable patterns, the pre-defined reference patch cannot locate the most similar position because many positions of the similar patterns will have indistinguishable similarities with the reference patch. Third, if the entire image is the reference, the area-based method will suffer from high computational complexity.

Compared to the area-based methods, feature-based methods are more robust to complex image distortions. They also cost less computations than the full image correlation-like methods. Conventional feature-based registration algorithms mainly include five steps. First, interest points are detected at distinctive locations in both the reference and the target images. The interest points can be corners[17], blobs[18], saddles[15], or features found by different types of keypoint detectors. Second, a certain region around every interest point is represented by a feature descriptor – typically an $n \times 1$ vector (e.g. SIFT[18], SURF[12], BRIEF[14], Saddle[15]). Third, the feature vectors are matched between the reference and the target images based on the distance of the feature vectors (e.g., Mahalanobis distance, spectral angular distance (SAD), and Euclidean distance). Fourth, a transformation model can be estimated based on the point-to-point matching result. Finally, the transformation model is implemented on the target image. Moreover, the non-integer coordinates of the transformed target image can be interpolated appropriately.

However, the feature-based methods still cannot solve the second limitation of the area-based methods. They also suffer from the lack of accuracy if the reference and the target images have too many similar and deformable patterns.

In [8], an correlation-based matching algorithm was proposed. the authors demonstrated that the inclusion of periodicity and constellation information in registration and matching can significantly improve the accuracy and robustness of the image registration. Furthermore, it can facilitate the metrology and inspection of a manufacturing process. Inspired by the periodicity and constellation registration algorithm, the registration accuracy and speed could be improved by taking advantage of the continuous spatiotemporal information between consecutive images and the speed stability of the moving target.

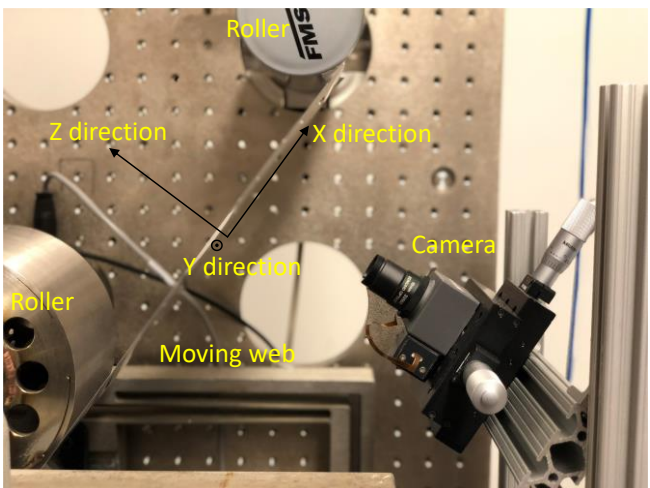

**FIGURE 1.** Experimental setup.

However, few of the feature-based methods used the coincident temporal and spatial information between every two consecutive images in an image sequence, which usually occurs in the continuous registration tasks (e.g., roll-to-roll printing systems [19]). In a typical roll-to-roll flexible electronic manufacturing process (see FIGURE 1), a fixed camera is generally used to inspect the printing quality. The images of flexible printing patterns will be acquired by the camera when the products move in one direction ($x$ direction in FIGURE 1). Given a sufficient frame rate for the product inspection, there will be many similar features in the overlapping areas between consecutive image pairs. These features can be duplicate printing patterns on the moving web. In these situations, the performance of a conventional feature-based registration algorithm is weak as these algorithms calculate the registration of each pair of consecutive images independently regardless of the smooth motion of the moving target and continuous spatiotemporal information in the consecutive image sequence, and thus cannot discriminate similar printing patterns.

Therefore, we propose a feature-based closed-loop feedback registration algorithm that will include the continuity of consecutive images and smoothness of target motion. Taking advantage of the preliminary spatiotemporal information of the consecutive image sequence, the first novelty of the proposed algorithm is that we generate a coarse translation model to guide the feature matching process. Instead of using the statistical model-based outlier removal methods, we revise the distance-based feature matching algorithm by generating the coarse translation model to eliminate the global search for all feature vectors spanning the entirety of the image. This approach can expedite the feature matching by reducing the computation complexity from $O(n^2)$ to $O(n)$, while improving the correspondence matching accuracy by including the estimation of a priori global moving.

The second novelty of the proposed algorithm is that we create closed-loop feedback to improve the accuracy of the coarse translation model. For consecutive images acquired by a camera of fixed pose (e.g., in a roll-to-roll printing system in FIGURE 1), the transformation model of the current image pair will be similar to the transformation model of the next image pair because of the smoothness and continuity of the motion of the imaged target given a fixed frame rate of the camera. Therefore, the optimized transformation model of the current image pair could be fed back into the coarse model generator of the next image pair. Consequently, this coarse-to-fine closed-loop feedback registration method can expedite the matching computation and serves to improve the matching accuracy and robustness.

The implementation of our proposed algorithm consists of three steps. First, we implement a conventional feature detector and descriptor (SIFT, SURF, and ORB are included in the experimental results) to find and represent the interest points in all images. Second, we implement the closed-loop feedback algorithm to match the interest points between each pair of the reference and the target images in a consecutive image sequence. Third, we stitch the series of the consecutive images using both direct linear transformation (DLT) and As-Projective-As-Possible (APAP) [20] transformation. APAP stitching can accommodate the deviations of the input data from the idealized conditions.

The remainder of the paper is organized as follows: Section II surveys related work of the registration task. Section III introduces our proposed registration algorithm. Experiment results are presented in Section IV. Conclusions and future work are presented in Section V and Section VI.

## II. Related work

The related work of area-based image registration methods is reviewed first in this section. Then, the related work of feature-based methods is reviewed according to the five steps of the image registration process.

Regarding the area-based registration methods, digital image correlation (DIC) has been widely used in various fields because of its ability to provide whole field deformation measurements and its convenience in terms of experimental preparation and data processing. In [21], the two iterative sub-pixel DIC registration algorithms, the inverse compositional Gauss-Newton (IC-GN) algorithm and the forward additive Newton-Raphson (FANR) algorithm, are fully analysed and evaluated. The authors claim that these two algorithms are the new standard of sub-pixel DIC. However, all these DIC-based methods [8], [10], [21] need a good pre-defined region of interest (ROI) patch and are not able to solve the complex transformations other than the translation. In [10], Kalman filter is used to predict the location and rotation of each pixel from current to the next deformed image in an image sequence. Then, the prediction is used as an initial guess of the correlation analysis. This method is well suited for applications where different parts of the image correspond to different motions as for example in the case of tracking multiple objects in a sequence of images. Nevertheless, the initial guess is limited to the translational and rotational transformations.

Regarding the first step of feature-based registration, many researchers have optimized interest point detection methods for various situations. In [17], a gradient-based method was adopted to find corners. The corners, named Harris corners, are detected by the second-moment matrix or second-order auto-correlation matrix. However, it is very sensitive to the variation of image scales. In [22], the Harris corner detector was improved by optimizing the feature selection process. Furthermore, the Harris-Laplace corner detector was proposed in [23], which is scale, rotation, and translation invariant. Interest points are extracted by the Harris corner detector at multiple scales of the Laplacian function and are selected at the maxima over the scales. In addition to corners, blob feature detectors are also adopted in many applications. The Laplacian of Gaussians (LoG) was applied to detect the blob-like features, which can be approximated by using the difference-of-Gaussian (DoG) approach for greater efficiency. In [11], Lowe introduced the SIFT detector to find blob features by DoG. Similar to SIFT, in [12], the SURF detector was introduced to find both corner and blob features. In [15], the Saddle detector was introduced to find the saddle-like features. The saddle condition was verified efficiently by intensity comparisons on two concentric rings with certain geometric constraints. In [24], edges were detected from images by a modified Canny operator, and line segments were then extracted from these edges. In addition, template-based feature detectors have been proposed in the literature [25]–[28]. Template-based methods are more comfortable for implementation, but variant to changes in scale.

Regarding the second step of the feature-based registration task, researchers designed various descriptors to keep the feature vectors distinctive and robust to many irregularities including, noise, detection errors, and geometric and photometric deformations. SIFT [15] is one of the most widely used gradient-based feature descriptors and is suitable for image matching and registration of optical images [29] or multi-spectral images [30]. SIFT descriptor constructs a gradient histogram in a local region to account for each feature. A 128-dimension vector is then created by assigning a Gaussian weighting function to each detected point. SURF [12] is another popular gradient-based local feature descriptor that uses the Haar Wavelet response for assigning an orientation histogram to represent features. SURF also improves the speed of the feature detection and description in SIFT. Similar to SURF, other researchers built upon SIFT to extend its advantages. In [31], PCA-SIFT was introduced based on normalized gradient patches and reduced window size ($41 \times 41$) to reduce the size of the descriptor. In [32], A-SIFT was proposed for affine transformed images. An n-SIFT algorithm [33] was proposed to change the unique SIFT algorithm by representing features in multiple dimensions. These multi-dimensional features can then be extracted and matched with 3D and 4D images. In [34], the SAR-SIFT algorithm including both detector and descriptor was proposed for synthetic aperture radar (SAR) images. The descriptor, named Circular-SIFT, is computed on a log-polar grid of nine sectors without using Gaussian weighting of the gradient magnitudes for computing histograms. Another gradient-based descriptor proposed in [35] used the distribution of normalized Laplacian spectra to characterize feature points. The robustness of the spectral matching methods against positional jitters and outliers was improved by the spectral representations of the point patterns. In [36], a feature descriptor called LDAHash was introduced which reduces the SIFT descriptor size by representing them with binary strings ultimately serving to reduce computational complexity. Then, the faster hamming distance, rather than Euclidean distance, was used for feature matching. BRIEF [14] and BRISK [27] use binary strings as the descriptors, taking advantage of the hamming distance metric. BRIEF is a binary string descriptor that obtains individual bits by comparing the intensities of paired points in the feature patch. In BRISK, a sampling pattern is applied to the neighbourhood of each detected keypoint to retrieve gray values, which are then used to define the orientation. The oriented sampling patterns provide the descriptor and make BRISK invariant to scale and rotation. In [37], partial intensity invariant feature descriptor (PIIFD) was introduced which is rotation invariant and partially invariant to affine and viewpoint changes.

Regarding the third step of the feature-based registration task, researchers have optimized other distance measurement metrics. The $m_p$-dissimilarity measure had been recently proposed in [38], [39] which was suitable to measure the data dependency similarity. Unlike $l_p$-norm distance (Euclidean being $l_2$-norm), $m_p$-dissimilarity considered the relative positions of the two vectors with respect to the remaining data. Furthermore, instead of one-to-one correspondence, some probability-based point set registration methods are proposed by representing the point sets as Gaussian mixture models (GMMs). In [40], the signature quadratic form distance was derived to measure the distribution similarity between two GMMs which was robust to noise, outliers, missing partial structures, and initial misalignments. In [41], the asymmetric Gaussian representation method was adopted instead of GMMs. However, the transformation achieved by these methods is limited to rotation and translation.

Unfortunately, these feature matching methods produce many incorrect matches when the pair of images contain a large number of similar features, such as the repeated regular printed patterns (grid-based images) in manufacturing [8]. To eliminate the incorrect matching points in the third step of the registration task, statistical model-based methods using geometric constraints are commonly used. The most popular method is random sample consensus (RANSAC) [42], which has several extensions, such as MLESAC [43], IMPSAC [44], and PROSAC [45]. These methods adopt a hypothesize-and-verify approach that attempts to find the best parametric model under their defined regulations. Nevertheless, these statistical model-based methods increase running time and perform unstably due to random sampling. Another method to eliminate the incorrect matching points is to preserve the local connectivity of the feature points. In [46], a probability-based registration algorithm was proposed that explored connectivity information of the reference point set to constrain the transformation model between the reference point set and the target point set. The $k$-connected neighbours are used to preserve the local neighbourhood information. Then the nonrigid transformation is formulated as a probability-based energy optimization problem and solved by the expectation-maximization (EM) algorithm. Like the statistical model-based methods, slow running time is also inevitable using this method. To improve the robustness and accuracy of registration of the locally distorted images, coarse-to-fine registration methods were proposed. In [30], [47]–[49], a preregistration process was implemented first, then followed by some of the different fine-tuned methods to refine the registration accuracy. In [50], an Iterative Scale-Invariant Feature Transform (ISIFT)-based registration algorithm was proposed. The authors continuously updated the matched point pairs between the sensed image and the reference image until an optimized transformation model was found. However, this method is time-consuming because of the large iterative optimization involved. Moreover, the four mutual information-based similarity metrics, which are used to update the matching point pairs in every iteration, are not robust enough as they are greatly impacted by the intensity changes of the image pixels. Furthermore, some researchers attempted to take advantage of the prior information in their specific situations for accurate registration. In [8], a constellation matching algorithm was proposed that used the Normalized Cross-correlation (NCC). The author used the preliminary spatial and pattern information of the images to achieve an accurate matching result. However, the templates for detecting interest points are needed and the initial reference point should be accurately paired at the beginning of the matching algorithm.

Finally, given the optimized matching point pairs, an image stitching algorithm is commonly adopted. The most straightforward method is to align images by estimating the 2-D projective warps. Parameterized by the $3 \times 3$ homography, the 2-D projective warps are justified in [51, pp. 273–292] if the scene is planar or if the views differ only by rotation. In practice, these viewing conditions may not be fully satisfied as the projective model can wrongly characterize the warp and thus cause misalignments or ghosting effects. To solve this problem, the APAP stitching method was proposed in [20] by constructing better alignment functions. The strategy of this method utilizes the local projective models created based upon locations of the images, named APAP, rather than using the global projective transformation model. This method can significantly reduce alignment errors while maintaining the overall geometric plausibility.

### III. Proposed algorithm

In this section, we will present the details of how the feedback registration system works. Intuitively, if we know the initial speed $v_0$ of the moving target (e.g., the moving web in FIGURE 1) and each image frame lacks significant deformation, for any frame at time $t$, $t > 0$, the moving distance between two consecutive frames can provide an approximate translation model of the image sequence as a prior registration at time $t$. Furthermore, the moving distance between the first frame and the second frame ($t = 0$ and $t = \Delta t$) can be obtained by $d_1 = v_0 \Delta t$, where $\Delta t$ represents the time interval of two consecutive frames, and the following moving distances $d_i$, $i > 1$, can refer to the previous moving distances, $d_{i-1}$. Then, the feature vectors could be matched by the regularizations provided through the prior registration, which differs from the global search for all vectors spanning the entirety of the image. Next, based on the previous matching result at time $t$, a revised translation model can be fed back into the next matching task at time $t + \Delta t$. FIGURE 2 shows the workflow of the system. Taking a series of frames as input, a constellation prediction model is created first; given the prior registration results, the point pattern matching (PPM) module optimizes the results and feeds the results back to the constellation prediction module for accurate registration results. In the following, we will detail each step of our registration algorithm.

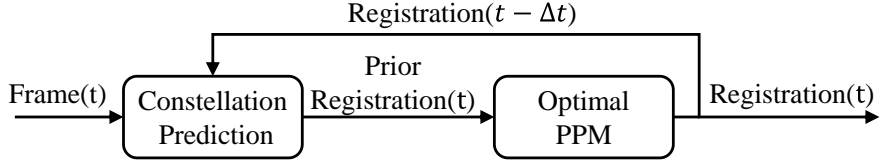

**FIGURE 2.** Workflow of Closed-loop Feedback Registration System.

*A. 2-D translation estimation*

Assume there is a global 2-D projective transformation between every two consecutive images [51, pp. 273–292]. Given any two consecutive images, $image_{k-1}$ and $image_k$, for $1 \leq k \leq n$, in an image sequence $\{image_0, image_1, \cdots, image_n\}$, we can transform all the pixels in $image_k$ from its own coordinate system into the coordinate system of $image_{k-1}$. The image transformation can be represented as:

$$I_k^{k-1} = T_{proj}(k) \cdot I_k, \quad (1)$$

where $I_k^{k-1}$ and $I_k$ both represent $image_k$, yet respectively in the coordinates of $image_{k-1}$ and $image_k$. $T_{proj}(k)$ represents the 2-D projective transformation matrix between $image_{k-1}$ and $image_k$. Therefore, given $I_k$ in the coordinate system of $image_0$, $I_k^0$ can be represented by:

$$I_k^0 = \prod_{i=1}^{k} T_{proj}(i) \cdot I_k. \quad (2)$$

For a manufacturing process, like roll-to-roll printing in FIGURE 1, the product target is typically controlled to move unidirectionally. We define the moving direction to be the $x$ axis. As a result, there will be minimal displacements in the directions of the $y$ and $z$ axes. Some insignificant offsets in the directions of the $y$ and $z$ axes might be a result from the misalignment and vibration of the moving conveyor (e.g., moving web in FIGURE 1). Therefore, given the setting speed of the moving conveyor ($v$ ($m/s$)), the frame rate of the camera ($f$ ($1/s$)), and the pixel scale of the images ($d(m/pixel)$), we should be able to approximately calculate the pixel offsets along the $x$-axis in-between two consecutive images. FIGURE 3 shows an example of the image acquisition process. For any frame at time $t$, where $t > 0$, the moving distance (pixels) between two consecutive images, $\Delta x$, can be presented as:

$$\Delta x = \frac{v}{f \cdot d}. \quad (3)$$

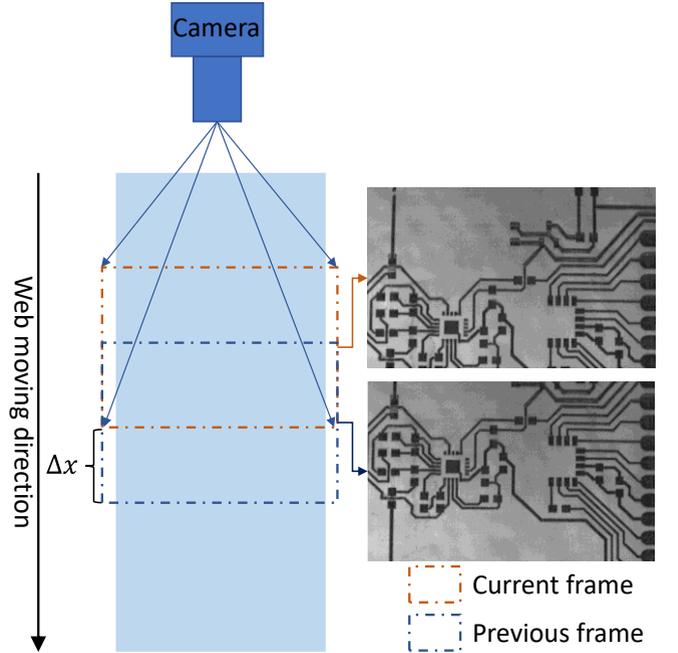

**FIGURE 3.** An example of the image acquisition process.

Given a sufficiently fast frame rate, we can estimate the global projective transformation model using a simpler translation model because the consecutive images in the short time slot will only have a significant offset in the positive $x$ axis as that axis is the principal component of moving direction. Therefore, $T_{proj}$ in (2) can be simplified by the translation transformation matrix $T_{trans}$, which can be expressed as:

$$T_{proj} = \begin{bmatrix} h_{11} & h_{12} & h_{13} \\ h_{21} & h_{22} & h_{23} \\ h_{31} & h_{32} & 1 \end{bmatrix} \longrightarrow T_{trans} = \begin{bmatrix} 1 & 0 & \Delta x \\ 0 & 1 & 0 \\ 0 & 0 & 1 \end{bmatrix}. \quad (4)$$

However, errors will be introduced in each estimation of the transformation due to the omission of offsets in $y$ and $z$

axes in (4). Substituting this simplified transformation matrix of (4) and its offset errors into (2), $I_k^0$ can be derived by:

$$I_k^0 = \prod_{i=1}^{k}[T_{trans}(i) + Error(i)] \cdot I_k, \quad (5)$$

where $T_{trans}(i)$ represents the $i^{th}$ estimated translation transformation matrix. $Error(i)$ represents the offset error in the $i^{th}$ estimation. Equation (5) indicates that the errors will be accumulated in each transformation estimation. We will present the error correction process using a feedback loop in the next part.

### B. Closed-loop feedback matching

The calculation in Section A involves the global 2-D transformation between images. A follow-up algorithm for seeking the 2-D transformation matrices in (2) and (5) usually starts with searching for the point correspondences between the reference image and the target image. However, conventional PPM algorithms identify the point correspondences only by measuring the similarity of the feature points extracted by the feature detector and descriptor while omitting the temporal continuity of the consecutive images. To improve the matching speed and accuracy of the temporal continuous moving images, we designed a closed-loop feedback matching algorithm below.

Conventional feature point matching methods generally compare the similarity of the feature vectors between the reference image and the target image. Without loss of generality, $image_{i-1}$ and $image_i$, where $1 \leq i \leq n$, can be used respectively as an example for the reference image and the target image taken from a consecutive image sequence $\{image_0, image_1, \cdots, image_n\}$. Assuming that $m$ and $n$ interest points are respectively detected in $image_{i-1}$ and $image_i$, the positions of the interest points are stored in the vectors $P_{i-1}$ and $P_i$. $V_{i-1}$ represents the matrix of the feature vectors for $image_{i-1}$ where each row of $V_{i-1}$ corresponds to each point position of $P_{i-1}$. Likewise, $V_i$ represents the matrix of feature vectors for $image_i$. The similarity between feature vectors in $V_{i-1}$ and $V_i$ will be computed using a certain distance metric (e.g., Euclidean distance) and used to identify the vector of greatest similarity in $V_i$ for the corresponding vector in $V_{i-1}$. If the distance between the feature vector of $V_{i-1}$ and its respective similar vector in $V_i$ is demonstrated to be below a certain threshold, they are considered to be a match. The matching process can be presented in FIGURE 4. As an example, $V_{i-1}(j_1)$ and $V_i(k_1)$ is the first match because $V_i(k_1)$, a feature vector in $V_i$, is the most similar feature vector to $V_{i-1}(j_1)$, and the distance between $V_{i-1}(j_1)$ and $V_i(k_1)$ is below the certain threshold. As the result of the matching process, there are $l$ matching point pairs in FIGURE 4.

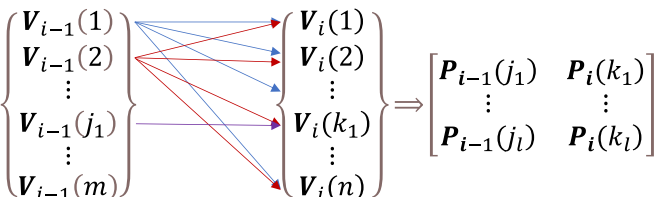

**FIGURE 4.** Conventional feature point matching process.

However, this matching method misses the global temporal and spatial continuity between the consecutive images. Moreover, it incurs a high computation complexity of $O(n^2)$, where $n$ represents the number of interest points. To improve the accuracy and speed of the PPM matching method, we take advantage of the translation model in Section A. Given the translation model $T_{trans}(i)$, we can easily derive the corresponding point position $P'_{i-1}$ for point position $P_{i-1}$, by:

$$P'_{i-1} = T^{-1}_{trans}(i) P_{i-1}. \quad (6)$$

where $P'_{i-1}$ represents the estimated locations of all the detected interest points in $image_{i-1}$ shown in the coordinate system of $image_i$. Equation (6) projects the detected interest points of $image_{i-1}$ into the subsequent coordinate system of $image_i$. Thereafter, in order to find the possibility of matching interest points between $image_i$ and $image_{i-1}$, only a small range around the positions of $P'_{i-1}$ needs to be considered. That is due to the coarse guideline for matching given by the estimated translation model. The interest points in $image_i$ outside the small range cannot be the correct matching points even though the feature vectors of these interest points are the most similar to any interest points in $image_{i-1}$. In summary, the matching point $P_i(k_1)$ in $image_i$ for an interest point $P_{i-1}(j_1)$ in $image_{i-1}$ can be found by (note that $V_{i-1}, V_i$ respectively represent feature vectors of $P_{i-1}, P_i$):

$$P_i(k_1) = \begin{cases} P_i(k) \in P_i(c): \\ V_i(k) = \mathbf{argmin}(\|V_i(k) - V_{i-1}(j_1)\|) \\ \cap (\|V_i(k) - V_{i-1}(j_1)\| \leq THR) \end{cases}, \quad (7)$$

$$P_i(c) = \begin{cases} P_i(c): P_i(c) \sqsubseteq P_i \cap \\ (P_i(c) \in Range(r) \text{ of } P'_{i-1}(j_1)) \end{cases}, \quad (8)$$

where $P_i(c)$ represents all candidate points in the range of $r$; $THR$ represents the distance threshold to match two feature vectors; $\mathbf{argmin}(\cdot)$ represents the feature vector that has the minimum distance.

FIGURE 5 illustrates the proposed matching algorithm. $P_{i-1}(1)$ represents an interest point found in $image_{i-1}$; $P'_{i-1}(1)$ represents the corresponding position of $P_{i-1}(1)$ in the coordinate system of $image_i$ after implementing the inverse translation transformation $T^{-1}_{trans}(i)$ for $P_{i-1}$; $r$ represents the small searching range which we will calculate subsequently. In the example shown in FIGURE 5, the corresponding matching point of $P_{i-1}(1)$ has only two possible candidates, $P_i(c_1)$ or $P_i(c_2)$ meaning only 2 iterations of comparison are needed. Therefore, the proposed matching algorithm considers the spatial information of the transformation so that the vast majority of the candidate points outside the searching range are removed. Furthermore, for each interest point in $image_{i-1}$, only a few comparative iterations are needed, signifying $O(n)$ complexity compared to the traditional $O(n^2)$.

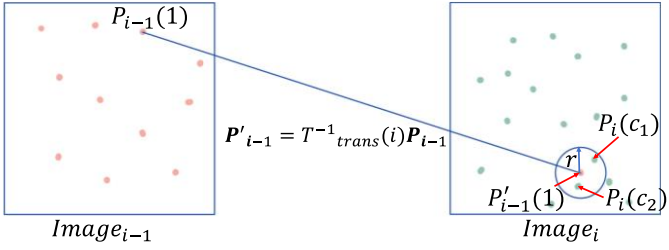

**FIGURE 5.** An example of the proposed matching process.

Given the matching point pairs $[P_{i-1}(J) \quad P_i(K)]$ (right side of FIGURE 4) from the algorithm in FIGURE 5, the average $\Delta x$ and $\Delta y$ of all the matching pairs can be calculated, creating a more accurate translation model to feedback to the constellation prediction module. Furthermore, in a roll-to-roll manufacturing process, the vibration in the $z$ axis is far more insignificant ($< 387$nm) compared to the main $x$ moving direction (mm scale), and the $y$ moving direction (mm scale), which is caused by the misalignment of the roller. Readers can refer to the supplementary material for more details. Therefore, the accumulating errors of Part A can be resolved by the feedback process. Meanwhile, a global projective transformation model could be easily derived by the direct linear transformation (DLT) method.

Moreover, when the global projective transformation model is inadequate, usually caused by the local deformation of the images, the APAP stitching algorithm [52] is adopted for the creation of panoramas. We will present both DLT and APAP experimental results in Section IV. FIGURE 6 shows the pseudo-code of the proposed registration algorithm.

### C. Calculation optimization

While the complexity of the proposed matching algorithm has been improved to $O(n)$ in Section B, its calculation can be further expedited for real-time applications. Instead of using loops for matching each element of $P_i(K)$ with $P_{i-1}(j)$ as shown in FIGURE 6, vectorization feature multiplication can be adopted for the matching distance calculation.

The sum of squared differences (SSD) of two vectors, e.g., $V_{i-1}(j)$ and $V_i(k)$ can be calculated by:

$$SSD_{jk} = \sum (V_{i-1}(j) - V_i(k))^2, \qquad (9)$$

where $\sum$ is the summation operator. However, calculating $SSD_{jk}$ for all combinational vector pairs in series using loops will be time-consuming. In the following, we attempt to

---

**Algorithm 1:** Closed-loop registration

**Input:** a sequence of images to be registered, $image_0$ to $image_k$, $\Delta x$ between the first two images

**Output:** The sequence of matching points pairs $\{M_1, M_2, \cdots, M_k\}$, the stitched $panorama$

1. Initialization: $\Delta y \leftarrow 0$, $i \leftarrow 1$
2. **while** $i \leq k$ **do**
3.     detect and represent features $V_{i-1}, P_{i-1}$ in $image_{i-1}$ and $V_i, P_i$ in $image_i$
4.     $j \leftarrow 1$
5.     **while** $j \leq length(V_{i-1})$ **do**
6.         use equation (7) to find $P'_{i-1}(j)$
7.         use equation (8) to find $P_i(c)$
8.         Match each element of $P_i(c)$ with $P_{i-1}(j)$
9.         **if** find a best match $P_i(k) \in P_i(c)$ **then**
10.             add pair $[P_{i-1}(j), P_i(k)]$ into $M_i$
11.         **end**
12.         $j \leftarrow j + 1$
13.     **end**
14.     save $M_i$
15.     update $\Delta x$ and $\Delta y$
16.     $i \leftarrow i + 1$
17. **end**
18. create $panorama$ based on $\{M_1, M_2, \cdots, M_k\}$

**FIGURE 6.** Closed-loop feedback registration algorithm for consecutive image sequences.

parallelize the calculation by vectorization. Mathematically, equation (9) can be expressed as:

$$SSD_{jk} = \sum V_{i-1}(j)^2 + \sum V_i(k)^2 - 2 \cdot V_{i-1}(j) \times V_i(k)^T, \quad (10)$$

where $V_i(k)^T$ denotes the transpose of $V_i(k)$. If we further expand (10) from one-to-one SSD into $m \times n$ SSDs (meaning there are $m$ features in $V_{i-1}$ and $n$ features in $V_i$), all the SSDs between $V_{i-1}$ and $V_i$ can be represented and calculated by:

$$SSD = \sum_{rows} V_{i-1}^2 \times [1\ 1\ \cdots\ 1]_{1 \times n} + [1\ 1\ \cdots\ 1]_{1 \times m}^T \times \sum_{columns} (V_i^T)^2 - 2 \cdot V_{i-1} \times V_i^T, \quad (11)$$

where the value of the $j^{th}$ row and $k^{th}$ column of $SSD$ is $SSD_{jk}$. Although the total floating-point operations (FLOPs) of (9) and (11) are similar, the running time will be significantly reduced because of the vectorization in (11). The reason is that computation in loops is inherently slower than matrix multiplication in hardware applications (e.g., multiple cores CPUs, and GPUs). Vectorization further enhances the $SSD$ calculation from $O(n)$ to $O(1)$. The running speed will be evaluated in Section IV. FIGURE 7 shows the comparison of the two methods. FIGURE 7 (a) shows the loop implementation of the feature matching algorithm. FIGURE 7 (b) shows the $SSD$ calculation process of (a). FIGURE 7 (c) shows the vectorization implementation of the feature matching algorithm. FIGURE 7 (d) shows the $SSD$ calculation process of (c). As an example, the dimensions of all the feature vectors are set to $1 \times 128$.

### D. Setting the small searching range

In Section B, we presented that our matching algorithm only searches for a small neighborhood range $r$ around $P'_{i-1}(j)$. An appropriate definition of range $r$ is needed. In this section, we define the small range $r$ in our roll-to-roll printing system as an example, by analyzing irregularities including the image deformation caused by the distortion of the flexible web, the web vibration, and the uncertainty of the web speed. When compared to other factors in a stable roll-to-roll system, the speed uncertainty in the moving direction plays the most important role in image deformation because in [53], the uncertainty of the moving speed can be $v \pm$

**Algorithm 2:** Loop Feature Matching Algorithm
**Input:** $V_{i-1}, P_{i-1}$ from $image_{i-1}$ and $V_i, P_i$ from $image_i$
**Output:** all the matching pairs $M_i$
1 Initialization: $l \leftarrow length(V_{i-1}), j \leftarrow 1\ M_i \leftarrow empty$
2 **while** $j \leq l$ **do**
3   // $V_i(c)$ comes from $P_i(c)$
  calculate $D_j$, the sum of absolute differences (SSD) between $V_{i-1}(j)$ and $V_i(c)$
4   find $D_j(k)$, the minimum SSD of $D_j$
5   **if** $D_j(k) \leq threshold$ **then**
6     add pair $[P_{i-1}(j), P_i(k)]$ into $M_i$
7   **end**
8 **end**

(a) Loop feature matching method

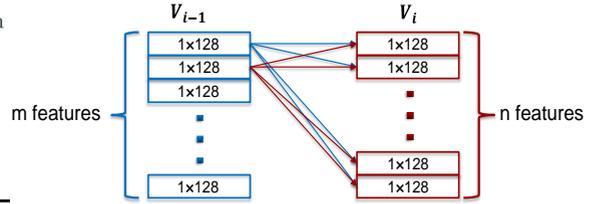

(b) SSD calculation process of loop matching

**Algorithm 3:** Vectorization Feature Matching Algorithm
**Input:** $V_{i-1}, P_{i-1}$ in $image_{i-1}$ and $V_i, P_i$ in $image_i$
**Output:** all the matching pairs $M_i$
1 Initialization: $l \leftarrow length(V_{i-1}), j \leftarrow 1\ M_i \leftarrow empty$
2 **while** $j \leq l$ **do**
3   find $P_i(c)$ for each $P_{i-1}(j)$
4   save it into $[j, c]$
5 **end**
6 implement equation (11) to find $SSD$
  // for $(V_i^T)^2$ and $V_{i-1} \times V_i^T$ parts, only calculate related values based on each $c$ calculated above, set other values into NaN
7 find $SSD_j^{min}$, the minimum values in each row of $SSD$
8 **while** $j \leq l$ **do**
9   **if** $SSD_j^{min} \leq threshold$ **then**
10     $k = index(SSD_j^{min})$
11     add pair $[P_{i-1}(j), P_i(k)]$ into $M_i$
12   **end**
13 **end**

(c) Vectorization feature matching method

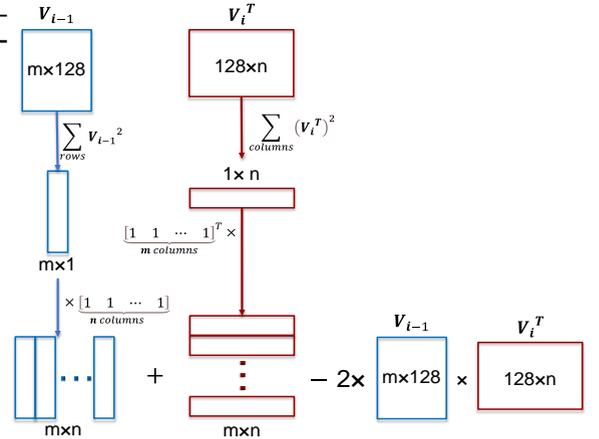

(d) SSD calculation process of vectorization matching

**FIGURE 7.** Comparisons of the loop and the vectorization matching implementations.

10%$v$ which is much larger compared to the vibration and distortion of the web. Therefore, $r$ should be defined mainly by the uncertainties in the moving direction ($x$ direction in FIGURE 1). In the following section, the small searching range $r$ is calculated based on the accuracy and robustness of the estimated translation model in A.

First, $r$ should be easily calculated if the speed range of the moving web is known. For example, in [53], the linear motion speed is controlled by PID controllers with variations less than 10%, so our range $r$, according to (3), should be defined as,

$$r = \left|\frac{v(1+10\%)-v}{f \cdot d}\right| \text{ or } \left|\frac{v-v(1-10\%)}{f \cdot d}\right| = \frac{\Delta x}{10}. \quad (12)$$

However, this range is calculated based on extremes. This range could be reduced in non-hypothetical scenarios. Assume the motion speed $v$ presents with truncated normal distribution [54] and the corresponding PDF shown in:

$$\psi(\mu, \sigma, a, b, x) = \begin{cases} \frac{\phi(\mu, \sigma^2)}{\Phi(\mu, \sigma^2; b) - \Phi(\mu, \sigma^2; a)} & \text{if } a < x < b \\ 0 & \text{otherwise} \end{cases}. \quad (13)$$

For the application of our roll-to-roll printing system, the bounds $a$ and $b$ are set at $0.9v$ and $1.1v$ respectively, while $\mu$ and $\sigma$ represent the mean and standard deviation of the motion speed $v$. In most engineering applications, covering 95% of the data should be sufficient, which means a smaller range $r$ could be derived according to certain web speed statistical parameters, $\mu$, and $\sigma$. Second, if the speed range of the moving target is unknown while the initial movement $\Delta x_0$ between first and second images is known, $r$ could be estimated experimentally. Third, if the initial movement $\Delta x_0$ is also unknown, an initial image pair must be registered using any of the classical methods. Then, $\Delta x_0$ could be calculated by this initial image pair.

## IV. Experimental results

In this section, we evaluate the performance of our closed-loop registration algorithm using both synthetic and real-world data. All these registration algorithms are evaluated in the MATLAB 2020a environment with Intel(R) Core (TM) i9-10980HK CPU @ 2.4 GHz laptop.

### A. Simulation data experiments

In this sub-section, we evaluate the performance of our closed-loop feedback registration algorithm with synthetic images. Consecutive repetitive and regular printed patterns are commonly encountered in real-world flexible electronics manufacturing due to the nature of printing, e.g., roll-to-roll printing. To simulate the images in the real-world production line of such repetitive and regular grid-based patterns, three synthetic image datasets are created. Each of the three datasets contains twenty images with a size of $400 \times 400$ pixels with their own unique basic grid patterns. The first dataset consists of squares with the side length of 41 pixels, the second dataset consists of circles with the radius of 20

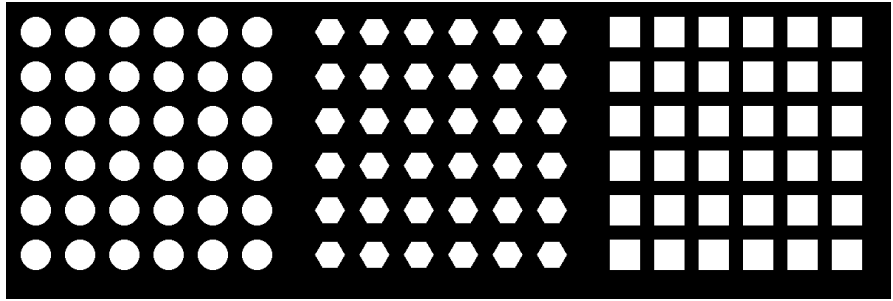
(a) The original grid-based images

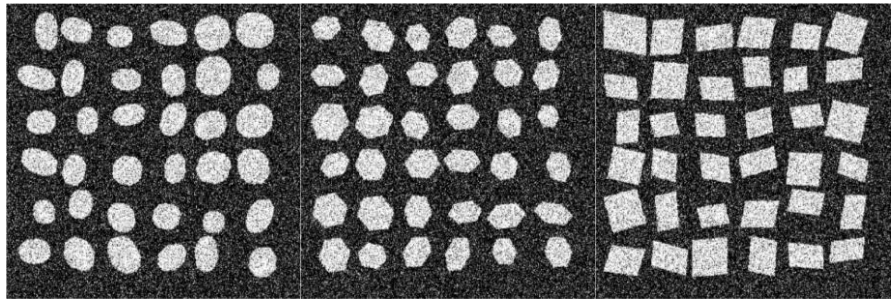
(b) The deformed grid-based images with distortion fact 3.0

**FIGURE 8.** An example of the three synthetic datasets. (a) without deformation. (b) with deformation.

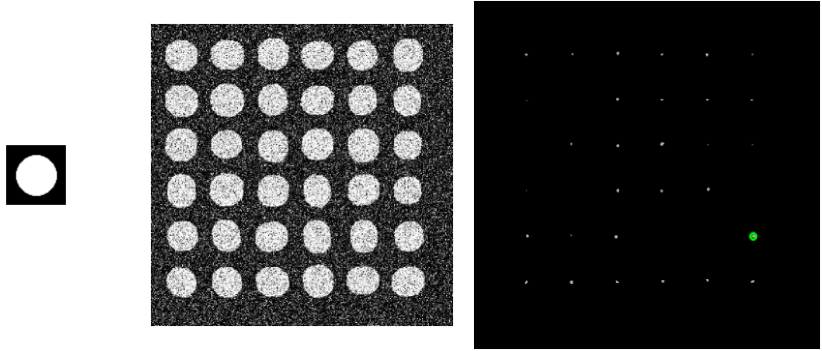

(a) correlation result of the first deformed image in the circle dataset

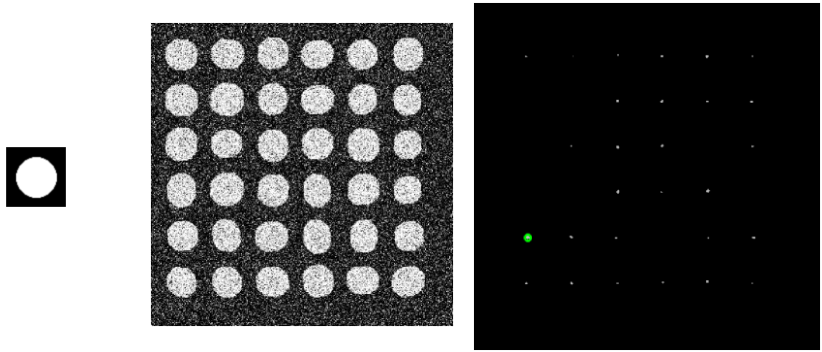

(b) correlation result of the second deformed image in the circle dataset

**FIGURE 9.** Correlation results of the deformed images in the circle dataset; from left to right are undeformed patch, distorted target image (distortion factor=1.0), normalized 2-D cross-correlation result. The white regions are all positions that the similarity is greater than 0.7, the green circles are the positions with the greatest similarity.

pixels, and the last dataset consists of hexagons with the radius of 20 pixels. All these patterns are deformed randomly by affine transformation locally. Meanwhile, each image is corrupted by Gaussian-blurring-filtering with a mask size of $7 \times 7$, salt and pepper noise of density of 0.05, and white Gaussian noise of variance of 0.02. We use a distortion factor to define the level of local affine deformation, i.e., the greater the distortion factor is, the broader the random translation, rotation, scaling, and shear ranges are (details can be found in [55]). FIGURE 8 shows an example of the three datasets. FIGURE 8 (a) shows the original three grid-based images. FIGURE 8 (b) shows the deformed images with the distortion factor set to 3.0. For each image in the three synthetic datasets, the center positions of all the deformed patterns are saved as the ground-truth when the image is created. We set the moving speed of each dataset to $dx = dy = 60 \ pixels/frame$, i.e., the distance between every two undistorted pattern's centers.

We first evaluate the performance of the conventional DIC algorithm. Taking one template of the undeformed patch, we use "normxcorr2" function in MATLAB to calculate the normalized 2-D cross-correlation between the template and the target image. However, as shown in FIGURE 9, there are many high similarity positions (greater than 0.7), and the greatest similarity position varies from different deformed target images. Therefore, conventional DIC algorithm cannot find the correct correspondences between each reference and targe image pair. The spatial-temporal information is required to further optimize the matching process. However, it will be still hard to implement DIC to the real-world images due to the difficulty of predefining the patches and the failure of handling transformations other than translation.

Consequently, three feature-based algorithms are implemented. Each of the SIFT, SURF, and ORB detectors and descriptors are used on a sequence of grid-based images to detect and describe the interest points. Then, these interest points are matched in various methods described below based on the moving speed. In the first matching method, we implement the lowest SSD metric directly. In the second method, the RANSAC outlier removal algorithm is applied to optimize the matching results in the first method. In the third method (if SIFT detector and descriptor are adopted), a new variant of SIFT, ISIFT[50] is implemented. ISIFT iteratively optimizes the results of RANSAC based on mutual information and shows higher registration accuracy than SIFT in remote sensing images. In the fourth method, the proposed closed-loop feedback matching method is implemented. An example of the matching results is shown in FIGURE 10. In FIGURE 10, all of the results are based on SIFT detector and descriptor. The results of SURF and ORB detectors and descriptors will be shown later. From FIGURE 10 (a)-(c), we can see both the traditional and the state-of-

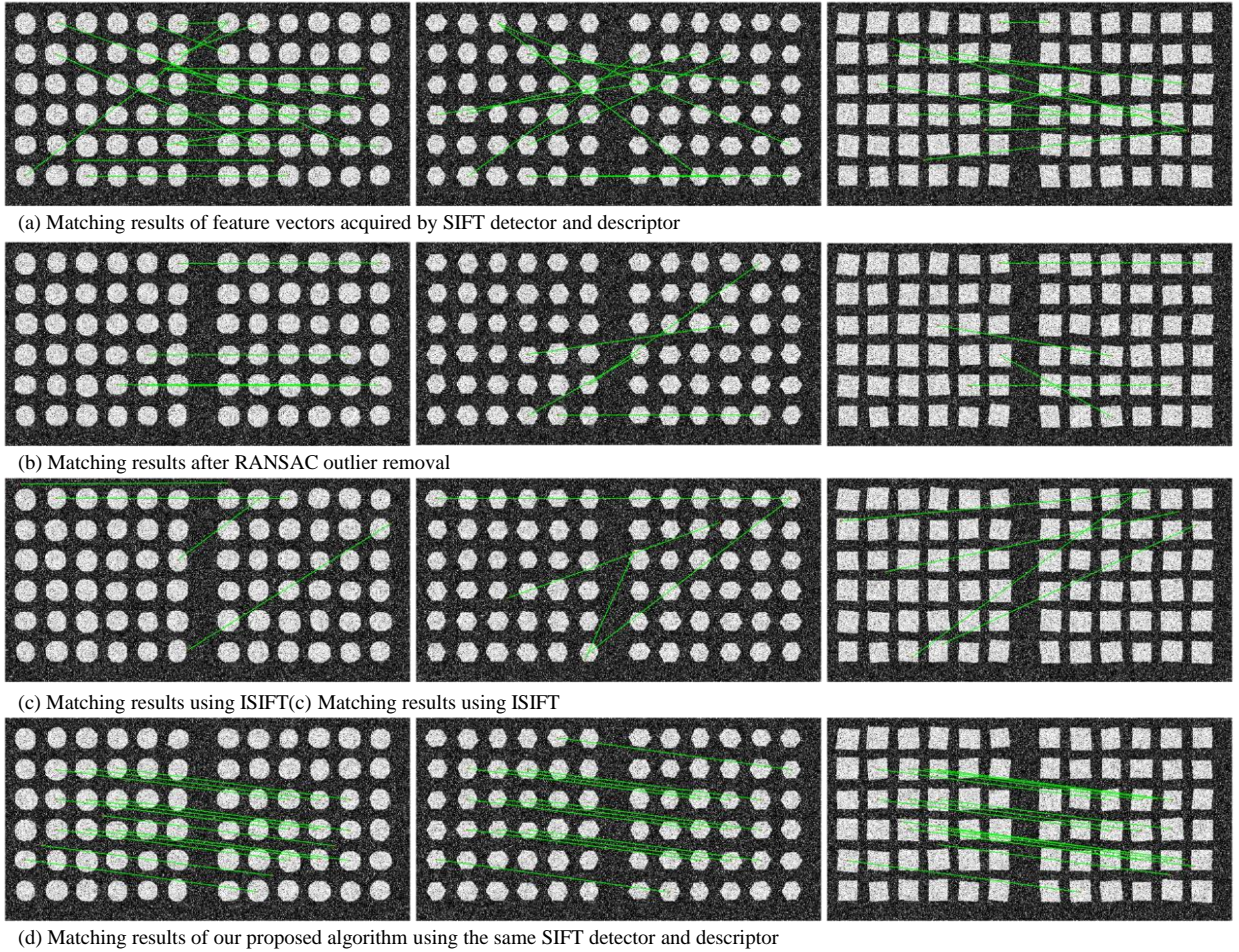

(a) Matching results of feature vectors acquired by SIFT detector and descriptor

(b) Matching results after RANSAC outlier removal

(c) Matching results using ISIFT(c) Matching results using ISIFT

(d) Matching results of our proposed algorithm using the same SIFT detector and descriptor

**FIGURE 10.** Comparisons of matching results based on SIFT detector and descriptor. (a) Matching with lowest SSD. (b) optimizing (a) with RANSAC outlier removal algorithm. (c) Matching with ISIFT algorithm. (d) Matching with the proposed closed-loop feedback matching algorithm.

the-art feature-based matching algorithms failed to find the correct correspondences between two grid-based images. The reason is that these algorithms ignore the smooth motion of the moving target and continuous spatiotemporal information in the consecutive image sequence, and thus cannot discriminate similar grid patterns. However, as shown in FIGURE 10 (d), taking advantage of the prior registration results, our proposed closed-loop feedback matching algorithm provides better accuracy and abundancy in the matching correspondences.

To further quantify the matching accuracy of the proposed algorithm, we first use the average Euclidean distances between the detected centers (we define the detected centers as the correspondences located inside the range of 5 pixels of the ground-truth centers) and the ground-truth centers to measure the detected center errors. Then, we also define the detected center ratio, which is the ratio of the number of the detected centers to the number of the ground-truth correspondences. Based on the moving speed ($dx = dy = 60\ pixels/frame$ in our experiment), the ground-truth centers of two consecutive images can be easily matched which can be defined as the ground-truth correspondences. Finally, to quantify the accuracy of all the detected matching correspondences (centers and non-centers), we compute the respective root mean squared error (RMSE) using:

$$RMSE(k) = \sqrt{\frac{1}{N}\sum_{i=1}^{N}\left\|T_{proj}(k) \cdot P_k(i) - P_{k-1}(i)\right\|^2}, \quad (14)$$

where $P_{k-1}(i)$ and $P_k(i)$ represent the coordinates of the $i^{th}$ matching point pair in $image_{k-1}$ and $image_k$. $T_{proj}(k)$ can be calculated with the ground-truth correspondences using the DLT method. Using (14), we transform the matching points in $image_k$ into the coordinate system of $image_{k-1}$ and then calculate the average $L^2$ distance of all the matching pairs in the coordinates of $image_{k-1}$. FIGURE 11 (a) shows an example of the ground-truth correspondences. FIGURE 11 (b) shows the matching result of the same image pair using the SURF detector and descriptor with our closed-loop feedback matching algorithm. The small black circles are the detected pattern centers of this image pair.

FIGURE 12 shows the quantitative errors from all three synthetic datasets. The horizontal axes are the distortion factors, the vertical axes on the left (blue color) are the errors of pixels, and the vertical axes on the right (red color) are the detected center ratios. Given different distortion factors, the blue lines are the average detected center errors, the green lines are the average RMSE of all the detected correspondences, and the red lines are the average detected center ratios. FIGURE 12 (a) shows the errors using the SIFT detector and descriptor and our matching algorithm, FIGURE 12 (b) shows the errors using SURF detector and descriptor and our matching algorithm, and FIGURE 12 (c) shows the errors using the ORB detector and descriptor and our matching algorithm.

First, the detected center errors of all three detectors and descriptors are less than 2 pixels. That proves our matching algorithm can locate the distorted pattern centers accurately. Second, the SURF detector finds the most centers and keeps the detected center ratio stable when the distortion factor even increases up to 2.0. The SIFT detector finds fewer centers than SURF and the detected center ratio decreases when the distortion factor increases. The pattern centers on the edges are partially missed which can be seen from FIGURE 11 (b). The reason is that the DoG detector in the SURF and SIFT algorithms is not stable on the borders of the images due to the zero-padding for the Gaussian blurring process. For the ORB detector, it finds more centers in the circle pattern dataset, fewer in the hexagon pattern dataset, and the least in the square pattern dataset. The reason is that the ORB detector is more sensitive to corners than the centers of the pattern in the hexagon and square pattern datasets. Last, compared to the other methods mentioned before, our closed-loop feedback matching algorithm successfully finds correct correspondences in the highly distorted (distortion factor up to 2.0) image sequences. Using the SIFT and SURF detectors and descriptors, the RMSEs are less than 3.5 pixels even though the distortion factor is up to 2.0. Using the ORB detector and descriptor, the RMSE is a little higher than SIFT and SURF, which is less than 8 pixels. The reason is that the SIFT and SURF descriptors are more robust to noise due to the longer feature vectors. However, the local distortions cannot be simply represented by a projective transformation matrix ($T_{proj}(k)$, calculated by the ground-truth centers using (14)). Therefore, the RMSEs of the closed-loop feedback matching algorithm are acceptable.

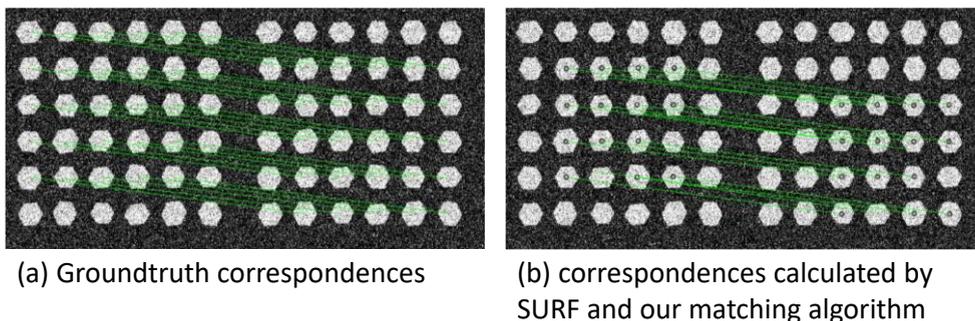

(a) Groundtruth correspondences    (b) correspondences calculated by SURF and our matching algorithm

**FIGURE 11.** Matching correspondences comparison. (a) Ground-truth correspondences calculated by the moving speed and the center positions of the patterns. (b) Matching results calculated by SURF detector and descriptor and our closed-loop feedback matching algorithm.

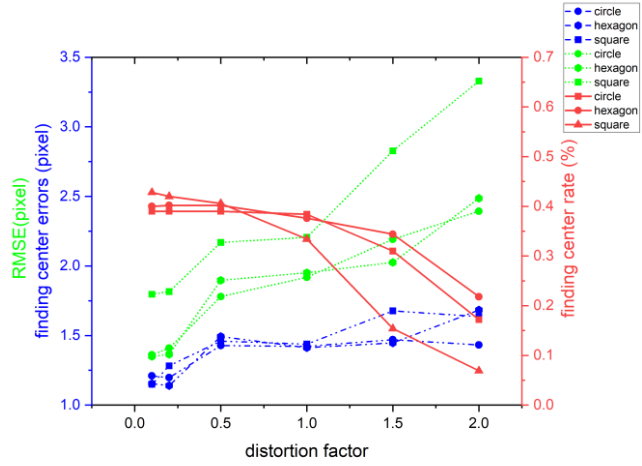
(a) Errors from SIFT

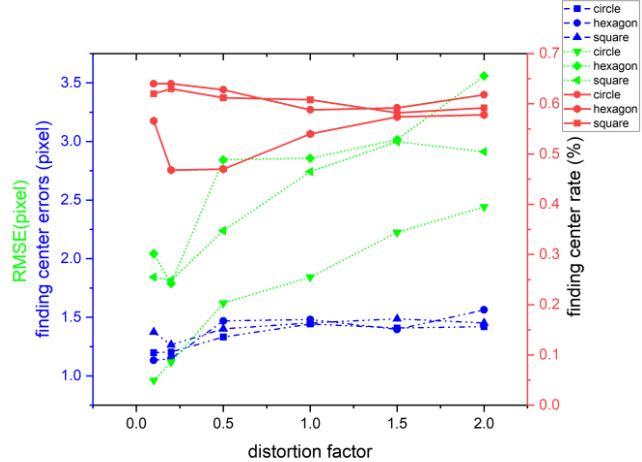
(b) Errors from SURF

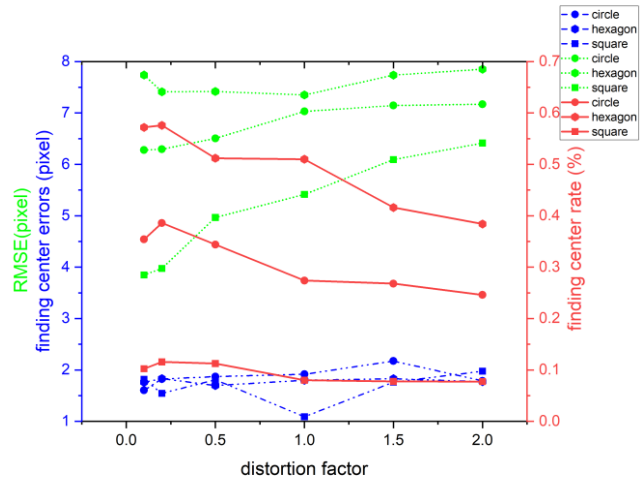
(c) Errors from ORB

**FIGURE 12.** Quantitative errors of our proposed algorithm on deformed grid-based images.

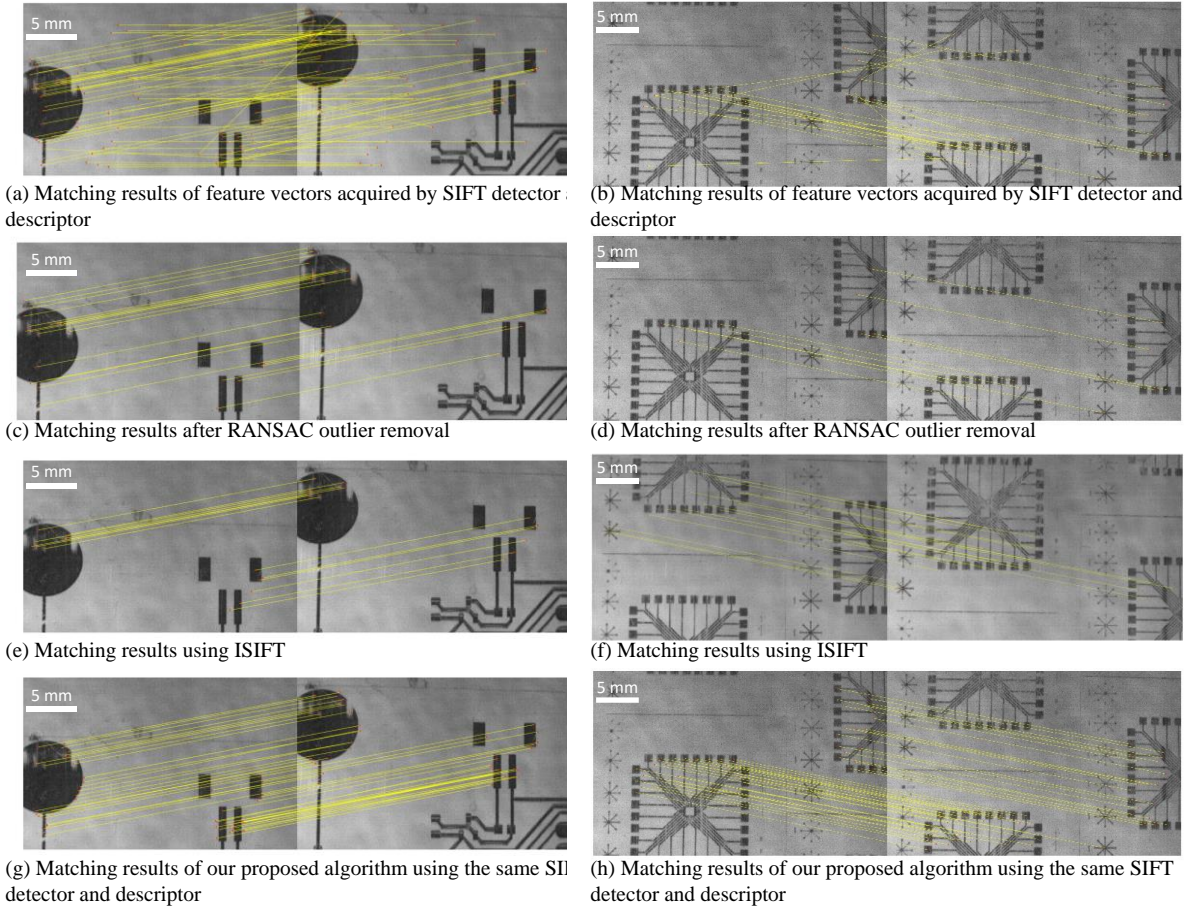

(a) Matching results of feature vectors acquired by SIFT detector and descriptor

(b) Matching results of feature vectors acquired by SIFT detector and descriptor

(c) Matching results after RANSAC outlier removal

(d) Matching results after RANSAC outlier removal

(e) Matching results using ISIFT

(f) Matching results using ISIFT

(g) Matching results of our proposed algorithm using the same SIFT detector and descriptor

(h) Matching results of our proposed algorithm using the same SIFT detector and descriptor

**FIGURE 13.** Matching examples in different methods. Left: R2R_Autofocus1; right: R2R_Autofocus2.

### B. Experiment setup on real-world moving flexible targets

To verify our proposed image registration algorithm in a real-world scenario, we set up experiments on a roll-to-roll print system (see FIGURE 1). The motion of the web along the $x$-axis can be controlled. A Pixelink PL-D721 autofocus camera with a 16 mm autofocus liquid lens was set up to inspect the circuits printed on the web. The printed patterns on the moving web are repetitive copper circuits. The motion speed of the web was set to 0.0127 m/s (0.5 inches/s). Since there are only minor vibrations in the $z$-axis compared to the motion in the $x$-axis, we set a small autofocus range between 38000-40000 dB to speed up the autofocus processing time [56]. Then, the focused images were captured at the frame rate of 2 frames per second (fps) in MATLAB 2020a environment with Intel(R) Core (TM) i9-9900X CPU @ 3.5 GHz desktop with an image size of $640 \times 1024$ pixels.

Two sequences of images are acquired by the Pixelink PL-D721 autofocus camera for the evaluation of the registration algorithms in the experiments. We called the datasets "R2R_Autofocus1" and "R2R_Autofocus2". R2R_Autofocus1 contains an entire circuit module. Each image of the dataset only contains a part of the module. R2R_Autofocus2 contains many duplicated circuit modules. Each image of the dataset can contain more than one module (see FIGURE 13 and FIGURE 15).

### C. Algorithm comparisons on real-world moving flexible targets

The performance evaluation experiments include the comparison of classic SIFT and RANSAC registration algorithms, the latest state-of-the-art ISIFT registration

TABLE I
Matching points comparisons of SIFT, SIFT with RANSAC, ISIFT, and the proposed algorithms

| datasets | Average percentages of matching points found from four algorithms | | | |
| --- | --- | --- | --- | --- |
| | SIFT | RANSAC | ISIFT | Ours |
| R2R_Autofocus1 | 100% | 39.59% | 10.97% | 65.69% |
| R2R_Autofocus2 | 100% | 33.43% | 9.71% | 60.32% |

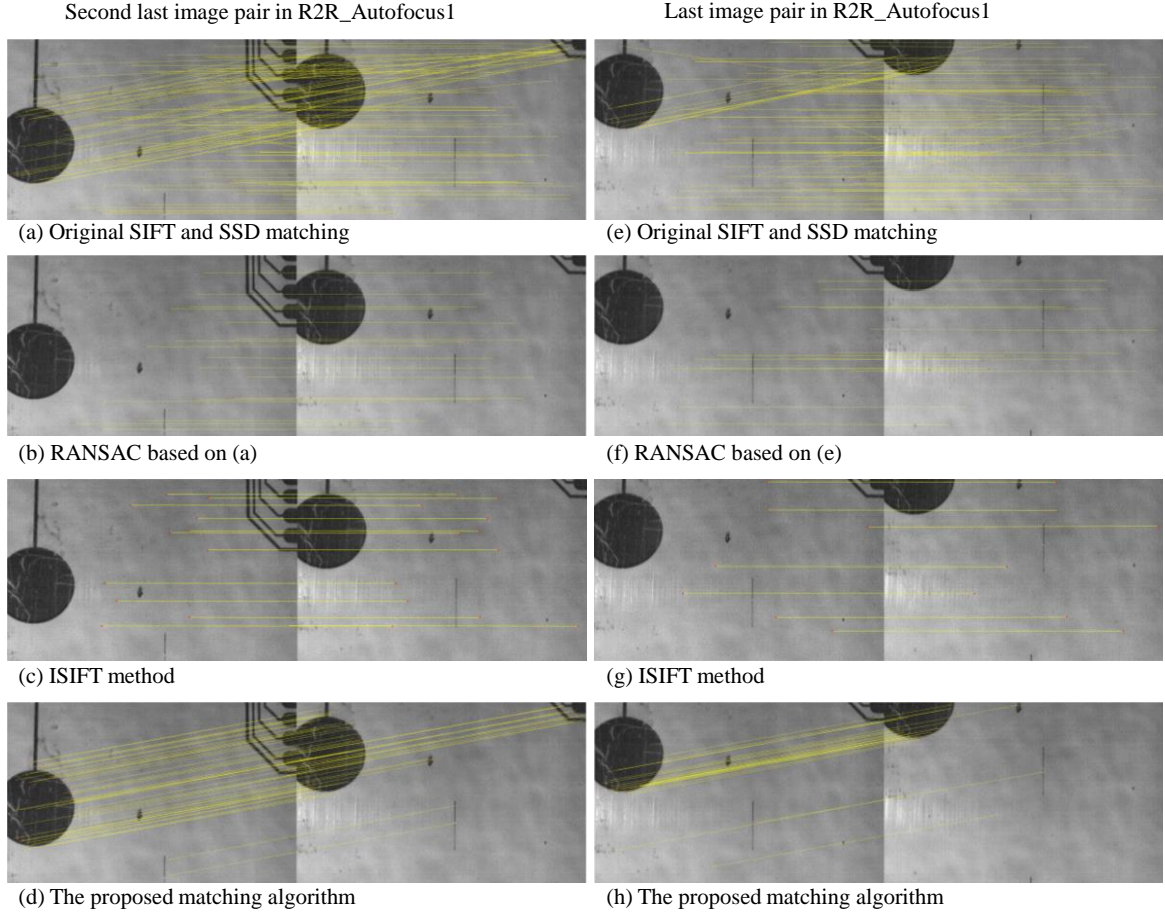

Second last image pair in R2R_Autofocus1     Last image pair in R2R_Autofocus1

(a) Original SIFT and SSD matching     (e) Original SIFT and SSD matching

(b) RANSAC based on (a)     (f) RANSAC based on (e)

(c) ISIFT method     (g) ISIFT method

(d) The proposed matching algorithm     (h) The proposed matching algorithm

**FIGURE 14.** Illustration of the matching results of the last two image pairs in R2R_Autofocus1.

algorithm, and our proposed algorithm.

Given a pair of images, the VLFeat library [57] is first used to detect and describe SIFT interest points. Then, these SIFT interest points are matched in various ways. In the first algorithm, the features from SIFT are directly matched using the lowest SSD metric. In the second algorithm, the RANSAC outlier removal algorithm is applied to optimize the matching in the first algorithm. In the third algorithm, ISIFT is implemented. In the fourth algorithm, the proposed closed-loop feedback matching method is implemented. FIGURE 13 shows the matching results of two image pairs in R2R_Autofocus1 (left) and R2R_Autofocus2 (right). FIGURE 13 (a) and (b) have many false positives because the two image pairs have many similar patterns spanning the entirety of the whole images. FIGURE 13 (c) and (d) remove the false positives but also remove some matching pairs. FIGURE 13 (e) and (f) find a similar number of matching pairs as (c) and (d), however, the results of ISIFT are not consistent in each experiment because of the iterative RANSAC process. FIGURE 13 (g) and (h) find far more matching pairs without any false positives which is beneficial to image stitching and overall will affect the inspection and quality control of the manufacturing line.

TABLE I shows the quantities of the average matching point pairs using all four matching methods in the R2R_Autofocus1 and R2R_Autofocus2 datasets. We set the basic SIFT and lowest SSD matching results as our baseline because the number of matching point pairs cannot be beyond the baseline if the threshold in FIGURE 7 is fixed. The results in TABLE I are consistent with the results of FIGURE 13.

Particularly, in the last two pairs of the consecutive images in R2R_Autofocus1, the matching results from SIFT with RANSAC and ISIFT are extremely inaccurate, as shown in (b), (c), (f), and (g) in FIGURE 14. The reason is that there are too many false positive matches in the original SIFT and lowest SSD matching algorithms in these two image pairs. RANSAC removes the correct matching point pairs as outliers because it tends to choose more quantities of the matching point pairs, which are false-positive matches in the scenarios of FIGURE 14. Furthermore, the ISIFT algorithm iteratively refines the RANSAC results based on the maxima of the mutual information between the image pairs, meaning it encounters the same problem as RANSAC. We perform 10 iterations of ISIFT experiments for matching the two image pairs in FIGURE 14 to try to eliminate the random nature of ISIFT as much as possible. Nevertheless, 9/10 of the experiments fail to find the correct matching point pairs in FIGURE 14. By contrast, as shown in FIGURE 14 (d) and (h), our proposed method finds the correct matching point pairs in both image pairs.

TABLE II
$\Delta x$ and $\Delta y$ in closed-loop feedback matching experiments

| datasets | Initial $\Delta x$ $\Delta y$ (pixels) | Average $\Delta x$ $\Delta y$ (pixels) | STD $\Delta x$ $\Delta y$ (pixels) |
|---|---|---|---|
| R2R_Autofocus1 | $\Delta x = -200$<br>$\Delta y = 0$ | $Mean(\Delta x) = -191.07$<br>$Mean(\Delta y) = 4.00$ | $STD(\Delta x) = 3.42$<br>$STD(\Delta y) = 1.66$ |
| R2R_Autofocus2 | $\Delta x = 200$<br>$\Delta y = 0$ | $Mean(\Delta x) = 196.49$<br>$Mean(\Delta y) = -3.81$ | $STD(\Delta x) = 2.09$<br>$STD(\Delta y) = 0.73$ |

TABLE III
RMSE of the matching results

| R2R_Autofocus1 | DLT RANSAC | DLT Ours | APAP RANSAC | APAP Ours |
|---|---|---|---|---|
| 3-4 | 3.8358 | 3.1808 | 3.8640 | 3.1045 |
| 4-5 | 3.7027 | 2.5773 | 3.6501 | 2.5767 |
| 5-6 | 2.4984 | 2.4012 | 2.5543 | 2.5368 |
| 6-7 | 2.7962 | 2.0037 | 2.8162 | 2.0343 |
| 7-8 | 4.4075 | 2.2133 | 4.4666 | 2.1878 |

TABLE II shows the initial, average, and standard deviation of $\Delta x$ and $\Delta y$ of the coarse translation models in both datasets. Using (3) and (4), we get the initial $\Delta x = -200$, $\Delta y = 0$ pixels in the R2R_Autofocus1 dataset and $\Delta x = 200$, $\Delta y = 0$ pixels in the R2R_Autofocus2 dataset. Then, the updated $\Delta x$ and $\Delta y$ continue to feed back to the next consecutive registration tasks. Using (12), (13), and the standard deviation of the web moving speed in [53] (refer to FIGURE 1), we get the searching range $r = 18$ pixels as default in the experiments. Then according to TABLE II, we find that the speed of the web is more stable than expected. Therefore, we further adjusted the small searching range to 10 pixels to accelerate the algorithm for real-time inspection.

FIGURE 15 shows the panoramas of R2R_Autofocus1. In the experiments, the SIFT with RANSAC and ISIFT methods have similar matching results, but the SIFT with RANSAC method is faster than the SIFT with ISIFT method due to the exclusion of the iterative RANSAC process. Therefore, to evaluate the stitching performance, we employ both DLT and APAP stitching algorithms based on the matching results from the SIFT with RANSAC method and the proposed method. FIGURE 15 (a) shows the panorama stitching results using the DLT stitching algorithm. SIFT with RANSAC matching results are used on the left side of FIGURE 15 (a). The proposed matching results are used on the right side of FIGURE 15 (a). FIGURE 15 (b) shows the panorama

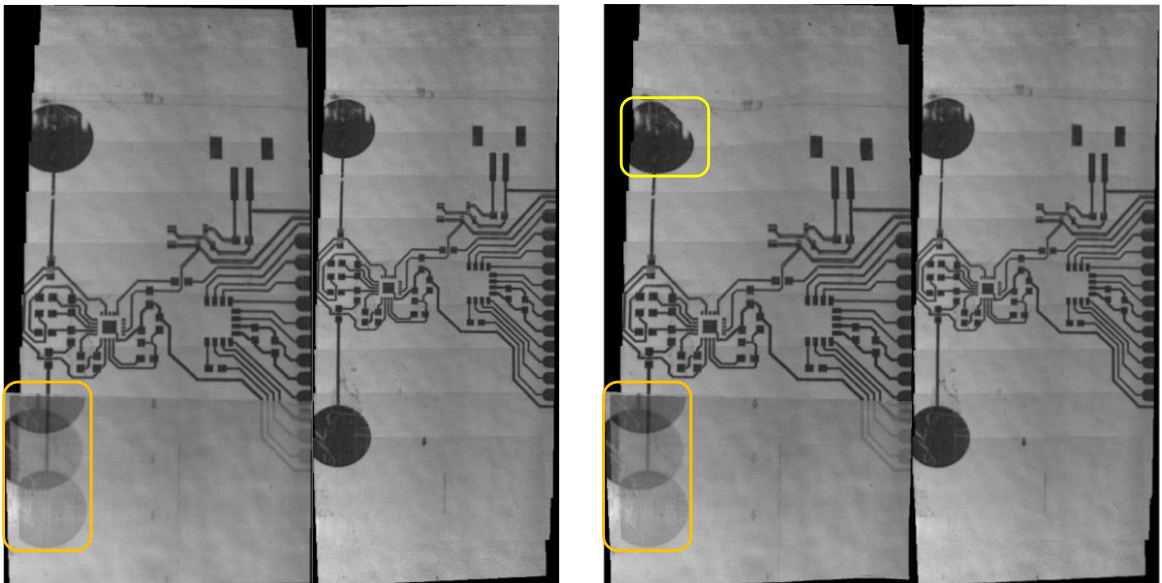

(a) Panoramas creation by DLT stitching.
Left: SIFT&RANSAC; right: our algorithm.

(b) Panoramas creation by APAP stitching.
Left: SIFT&RANSAC; right: our algorithm.

**FIGURE 15.** The panoramas of R2R_Autofocus1 created by SIFT with RANSAC and the proposed algorithm.

TABLE IV
Average running time comparisons of SIFT, SURF, and ORB detectors and descriptors with different matching algorithms, as well as ISIFT algorithm

| | Running time (seconds) | | | |
|---|---|---|---|---|
| | Without vectorization | | With vectorization | |
| | R2R_Autofocus1 | R2R_Autofocus2 | R2R_Autofocus1 | R2R_Autofocus2 |
| SIFT + Lowest SSD | 1.58 | 1.32 | 0.26 | 0.22 |
| SIFT + Lowest SSD + RANSAC | 2.21 | 1.51 | 0.58 | 0.51 |
| SIFT + Ours | 0.77 | 0.50 | 0.25 | 0.22 |
| SURF + Lowest SSD | 0.41 | 0.38 | 0.07 | 0.07 |
| SURF + Lowest SSD + RANSAC | 1.16 | 0.81 | 0.52 | 0.45 |
| SURF + Ours | 0.26 | 0.22 | 0.07 | 0.07 |
| ORB + Lowest SSD | 25.82 | 29.97 | 0.65 | 1.01 |
| ORB + Lowest SSD + RANSAC | 27.01 | 34.64 | 1.57 | 2.46 |
| ORB + Ours | 5.87 | 6.82 | 0.62 | 0.88 |
| ISIFT | 143.69 | 66.85 | 143.83 | 63.37 |

stitching results using the APAP stitching algorithm. SIFT with RANSAC matching results are used on the left side of FIGURE 15 (b). The proposed matching results are used on the right side of FIGURE 15 (b). Bundle adjustment [58] is applied during stitching. In each of the matching results on the left side of both FIGURE 15 (a) and (b), the matching correspondences on the bottom section (orange regions in FIGURE 15) have more misalignments and ghosting effects because of the mismatches in the registration processes. As shown in FIGURE 14, the SIFT with RANSAC method gave the incorrect matching results in the last two image pairs. Incorrect global projective transformation matrices are created during DLT stitching, resulting in the ghosting effects seen in the left side of FIGURE 15 (a) and (b). Additionally, despite its attempt to stitch the image pairs both locally and globally, the APAP stitching still cannot resolve the problem of producing wildly inaccurate global projective transformation matrices.

Furthermore, in FIGURE 15 (b), the top-left region highlighted in yellow has an obvious deformation when the APAP stitching method is implemented. The deformation comes from the local transformation matrix that the APAP algorithm finds in order to satisfy the matching points around this top-left region, resulting in false point correspondences in this region. The reason is that the APAP stitching algorithm tries to find many local transformation matrices according to the local matching correspondence results in certain regions. If an arbitrary region has too many false positive matching point pairs, the local transformation matrix of this region will be inaccurate, causing deformations during stitching.

However, on the right side of both FIGURE 15 (a) and (b), the stitching results are much better when compared to the left. The panoramas on the right side lack any ghosting effects or obvious deformations. The results prove that the proposed registration algorithm finds more accurate matching correspondences than the SIFT with RANSAC method.

To further quantify the matching accuracy, we compute the RMSEs of 'SIFT + RANSAC+DLT+APAP' and 'Our registration algorithm + DLT + APAP'. To make a fair comparison, we only choose the image pairs that have more than 30 matched points and are not completely inaccurate matches. In DLT stitching, the global projective homographic transformation $T_{proj}(k)$ is implemented on all the matching points in $image_k$. Then the RMSE is calculated using (14). Similarly, for the APAP stitching method, we do the same RMSE calculation using the local projective homographic transformation $h_*$ [20] for each matching point instead of $T_{proj}(k)$ for all the matching points in DLT method. TABLE III shows the RMSE results for the chosen image pairs in the R2R_Autofocus1 dataset. The proposed matching results have less RMSE than the SIFT & RANSAC method which proves our closed-loop registration algorithm performs better in this experiment.

### D. Running time evaluation on real-world moving flexible targets

In addition to accuracy, the fast speed of the registration method is essential for the real-time inspection and monitoring of manufacturing processes. In Section C, we propose a vectorization method to expedite the proposed matching algorithm. To evaluate the efficiency of the proposed algorithm, SIFT, SURF, and ORB detectors and descriptors are implemented for the registration of both R2R_Autofocus1 and R2R_Autofocus2 datasets for comparison. TABLE IV shows the average processing time per frame in different matching algorithms with SIFT, SURF, ORB detectors and descriptors, as well as the ISIFT method. The average processing time of the non-vectorization implementation of each method is demonstrated on the left-side of TABLE IV; the average processing time of vectorization implementation of each method is demonstrated on the right-side. The non-vectorization results on the left-side indicate the total FLOPs of each method. It is evident that our closed-loop feedback matching algorithm has the least FLOPs in all three detectors and descriptors. Particularly in the ORB detector and descriptor, the speed of our algorithm is five times faster than the lowest SSD matching method, and six times faster than the lowest SSD plus RANSAC outlier removal method. This is due to the fact that the ORB detector finds far more interest points than SIFT and SURF while our proposed algorithm removes many of these points as outliers efficiently. The vectorization results on the right-side show that all the registration methods are expedited from vectorization. In the SIFT and SURF detectors and descriptors, our algorithm is as fast as the lowest SSD matching method, yet much more accurate (see FIGURE 13 (a) versus (g), (b) versus (h), FIGURE 14 (a) versus (d), (e) versus (h)). In the ORB detector and descriptor, our algorithm is even faster than the lowest SSD matching method due to the large number of interest points. Furthermore, the time cost of the RANSAC outlier removal process dominates the total running time in the vectorization implementation because RANSAC cannot be expedited with vectorization. The ISIFT results show that the time cost from ISIFT is lengthy both with vectorization and without vectorization. The reason for this is that the iterative RANSAC process dominates the total running time in ISIFT. In summary, the experimental running time results show that the proposed algorithm is far more efficient than other registration algorithms.

## V. Conclusion

In this paper, a closed-loop feedback image registration algorithm is proposed. The algorithm involves two modules, namely the constellation prediction module and the PPM module. In the first module, a coarse translation transformation model is created using the spatiotemporal information of the consecutive image sequence. In the second module, the coarse translation transformation model is optimized to obtain an accurate point-to-point matching result. The accurate matching results keep feeding back into the first module to fine-tune the next coarse translation transformation model. By taking advantage of the spatiotemporal information of the consecutive image sequence, the proposed image registration algorithm demonstrates higher speed and accuracy than the state-of-the-art registration algorithms in matching point correspondences. The experiments on both simulation and real-world data show promising results for the registration of moving flexible targets. Our algorithm can accurately match more keypoint correspondences and achieve a better panorama than other registration algorithms. Additionally, the running complexity of the keypoint matching process is reduced from $O(n^2)$ to $O(n)$ and further accelerated up to 0.07 seconds per image pair using vectorization in MATLAB 2020a environment.

## VI. Future work

Translation estimation of the projective transformation model was employed for feedback loop registration in this paper. However, for different application scenarios, the hidden coarse models of transformation could vary. We will extend our estimation of the unidirectional transformation model to translation in both directions, scale, and mixed transformation models. Additionally, the proposed visual inspection system can be applied to the control of the roll-to-roll flexible electronics printing system. The speed and position of the moving web can be adjusted by the feedback data from the registration results to guarantee the quality of the products on the manufacturing line.

The Pixelink PL-D721 autofocus camera with a 16 mm autofocus liquid lens can only get as fast as 2 frames per second which is the speed bottleneck of the registration process. Advanced auto-focus algorithms[59] will be included in the future for real-time inline inspection.